\documentclass[runningheads]{llncs}
\usepackage{graphicx}
\usepackage{latexsym}
\usepackage[]{algorithm2e}
\usepackage{amsmath}
\usepackage{amssymb}
\usepackage{hyperref}

\begin{document}
\title{A generalized quadratic loss for SVM and Deep Neural Networks.\thanks{Supported by organization Universita' degli Studi "Ca' Foscari'' di Venezia.}}
%
%
\author{Filippo Portera\inst{1}\orcidID{0000-0002-2179-372X}}
\authorrunning{F. Portera}
%
\institute{Universita' degli Studi "Ca' Foscari'' di Venezia}
%
\maketitle              
\begin{abstract}
  We consider some supervised binary classification tasks and a regression task, whereas SVM and Deep Learning, at present, exhibit the best generalization performances.
  We extend the work \cite{ecai} on a generalized quadratic loss for learning problems that examines pattern correlations in order to concentrate the learning problem into input space regions where patterns are more densely distributed.
  From a shallow methods point of view (e.g.: SVM), since the following mathematical derivation of problem (9) in \cite{ecai} is incorrect, we restart from problem (8) in \cite{ecai} and we try to solve it with one procedure that iterates over the dual variables until the primal and dual objective functions converge. In addition we propose another algorithm that tries to solve the classification problem directly from the primal problem formulation. We make also use of Multiple Kernel Learning to improve generalization performances.
Moreover, we introduce for the first time a custom loss that takes in consideration pattern correlation for a shallow and a Deep Learning task. 
   We propose some pattern selection criteria and the results on 4 UCI data-sets for the SVM method. We also report the results on a larger binary classification data-set based on Twitter, again drawn from UCI, combined with shallow Learning Neural Networks, with and without the generalized quadratic loss.
At last, we test our loss with a Deep Neural Network within a larger regression task taken from UCI.
   We compare the results of our optimizers with the well known solver $\text{SVM}^{\text{light}}$ and with Keras Multi-Layers Neural Networks with standard losses and with a parameterized generalized quadratic loss, and we obtain comparable results \footnote{Code is available at: \href{https://osf.io/fbzsc/}{https://osf.io/fbzsc/}}.

\end{abstract}

\section{Introduction}

SVM and Neural Networks methods are widely used to solve binary classification, multi-class classification, regression tasks, $\dots$ .
In supervised binary classification learning tasks, SVM and Deep Learning methods are spread and represents the state-of-the-art in achieving best generalization performances.
The work of \cite{Joachims} and \cite{SMO} showed the potential different implementations of SVM, while there are different software to develop Deep Neural Network such as TensorFlow and PyTorch, to name a few.
Our goal is to improve the generalization performances of those algorithms, considering pattern correlations in the loss function.
In \cite{ecai} we proposed a generalized quadratic loss for SVM, but we mathematical development was erroneous. In details,
\cite{ecai} presents a step from eq. 8 and eq. 9 that is wrong, since we said that $(\vec{\alpha} + \vec{\lambda})' \mathbf{S}^{-1} (\vec{\alpha} + \vec{\lambda})$ is monotonically increasing, and this is not proved. We don't try to solve this problem, we restart from problem 8 and try to find methods to solve it in its dual and primal form.
Nevertheless the idea could be valid since if the matrix used to implement the loss is the identity matrix the loss reduces to the well know quadratic loss.
Here we develop the loss introduced in \cite{ecai} further, in the sense that we propose $2$ correct optimizers for the SVM setting and, perhaps more interesting, a custom loss that can be plugged-in into a Deep Learning framework.
In section \ref{works} we cover some related works about the problem we are studying.
In section \ref{loss_SVM_problem} we state the mathematical problem and we present the matrix $S$.
In section \ref{notation} we report some definitions for the proposed algorithms.
In subsection \ref{opt2} we describe the SMOS optimization technique.
In subsection \ref{opt3} we characterize the RTS optimization technique.
In subsection \ref{DLS} we elucidate the Deep Learning framework that exploits a loss function defined with the $S$ matrix.
In section \ref{results} we present the results obtained with 2 artificially generated data-sets, 4 binary classification data-sets for SVM, and 2 larger data-sets experiments carried on with Multiple Layers Neural Nets.
Finally, in section \ref{conclusions}, we draw some statements about the overall procedure and the results.

\section{Related works}\label{works}
We explore the use of a new loss with two different scopes: SVM and Neural Networks.
The canonical SVM model was first introduced in \cite{cortes_vapink}, and the losses used are a class of loss functions that doesn't take into consideration pattern correlations. Several optimizers have been proposed for this model and they can be found in \cite{Joachims}, \cite{SMO}, \cite{libsvm}, etc$\dots$, and almost all of them are based on the linear loss version of SVM.
On the other side, again, the losses used in Shallow and Deep Neural Networks are not considering pattern distribution.
For the Twitter sentiment analysis task there is a work \cite{Severyn} that propose a Deep Convolutional Neural Network approach and \cite{cliche-2017-bb} where they add also an attempt with LSTMs.
The YearPredictionMSD data-set has been studied in \cite{Lobato}, and in \cite{Lakshminarayanan}, to name a few.

\section{The modified loss SVM problem}\label{loss_SVM_problem}
In order to see if there is space for better generalization performances, 
we introduce a new loss as stated in \cite{ecai}.
For the rest of the paper we will use an $S$ matrix defined as:

\begin{equation}\label{S_definition}
  S_{i,j} = e^{- \gamma_S ||\vec{x_i} - \vec{x_j}  ||^2}
\end{equation}

which is a symmetric, positive semi-definite, and invertible matrix (if there aren't repeated patterns).

Therefore, reconsidering section 2 of \cite{ecai}, we obtain the dual problem:

\begin{equation}\label{dual_S}
  \vec{a}' \vec{1} - \frac{1}{2} \vec{\alpha}' \mathbf{Y} \mathbf{K} \mathbf{Y} \vec{\alpha} -\frac{1}{4 C} (\vec{\alpha} + \vec{\lambda})' \mathbf{S}^{-1} (\vec{\alpha} + \vec{\lambda})
\end{equation}

subject to the following constraints:
\begin{equation}\label{const_1}
\vec{\alpha}'\vec{y} = \vec{0}
\end{equation}
\begin{equation}
	i = {1,\dots, l}:
\end{equation}
\begin{equation}\label{const_2}
\vec{\alpha}_i \geq 0
\end{equation}
\begin{equation}\label{const_3}
\vec{\lambda}_i \geq 0
\end{equation}

\section{Notation}\label{notation}
In order to have a dual problem, $\mathbf{S}$ must be invertible.
If there are repeated patterns in the training set, $\mathbf{S}$ is not invertible. Thus we remove the repeated patterns from the training set.

Let: 
\begin{equation}
  i = {1,\dots,l},\ j = {1,\dots,l}:	
\end{equation}
\begin{equation}
  K(\vec{x}_i,\vec{x}_j) = e^{- \gamma_S ||\vec{x_i} - \vec{x_j}  ||^2}
\end{equation}
\begin{equation}
  f(\vec{x}_i) = \sum_{j=1}^{l} \vec{\alpha}_j \vec{y}_j K(\vec{x}_i, \vec{x}_j) + b
\end{equation}



\section{Algorithms}
In the following we propose 2 different algorithms to solve the primal problem.

\subsection{The SMOS optimization algorithm}\label{opt2}
With the same method reported in \cite{SMO-like}, we isolate the part of the dual function that depends on the updated variables:
\begin{equation}\label{SMOS_update_var_alpha_i}
  \vec{\alpha}_{i+1} \leftarrow \vec{\alpha}_i + \vec{\varepsilon}_i = \vec{\alpha}_i + \nu \vec{y}_i\\
\end{equation}
\begin{equation}\label{SMOS_update_var_alpha_j}
  \vec{\alpha}_{j+1} \leftarrow \vec{\alpha}_j + \vec{\varepsilon}_j = \vec{\alpha}_j - \nu \vec{y}_j\\
\end{equation}
\begin{equation}\label{SMOS_update_var_lambda_k}
  \vec{\lambda}_{k+1} \leftarrow \vec{\lambda}_k + \vec{\mu}_k
\end{equation}

where $\vec{\varepsilon}$ is a vector of dimension $l$ of all zeros, apart the $i$ and $j$ components that are, respectively, $\vec{\varepsilon}_i = \nu \vec{y}_i$ and $\vec{\varepsilon}_j = -\nu \vec{y}_j$.
While $\vec{\mu}$ is a vector of dimension $l$ of all zeros, apart the $k$ component that is equal to $\vec{\mu}_k$.


Omitting the derivation from the dual function of the optimized variables and deriving $D(\vec{\alpha} + \vec{\varepsilon}, \lambda + \vec{\mu})$ by $\nu$, and setting the partial derivative to $0$ in order to get the maximum for a fixed $\mu_k$, we obtain:


\begin{eqnarray*}
	\psi = \vec{y}_i - \vec{y}_j + \\
	+ \sum_{p=1}^{l} \vec{\alpha}_p  \vec{y}_p K(\vec{x}_p, \vec{x}_i)      
	- \sum_{p=1}^{l} \vec{\alpha}_p  \vec{y}_p K(\vec{x}_p, \vec{x}_j)   +  \\   
	- \frac{1}{2 C} (\sum_{p=1}^{l}  \vec{\alpha}_p \vec{y}_i S^{-1}[p,i] 
	- \sum_{p=1}^{l} \vec{\alpha}_p  \vec{y}_j S^{-1}[p,j] +                    \\
	+ \sum_{p=1}^{l} \vec{\lambda}_p \vec{y}_i S^{-1}[p,i]                    
	- \sum_{p=1}^{l} \vec{\lambda}_p \vec{y}_j S^{-1}[p,j] +                    \\
	+ \sum_{p=1}^{l} \vec{\mu}_p     \vec{y}_i S^{-1}[p,i]                     
	- \sum_{p=1}^{l} \vec{\mu}_p     \vec{y}_j S^{-1}[p,j] )
\end{eqnarray*}
and:
\begin{eqnarray*}
	\omega = K(\vec{x}_i, \vec{x}_i) - 2 K(\vec{x}_i, \vec{x}_j) + K(\vec{x}_j, \vec{x}_j)+\\
    + \frac{S^{-1}[i,i] -2 \vec{y}_i \vec{y}_j S^{-1}[i,j] + S^{-1}[j,j]}{2C}
\end{eqnarray*}

which implies:
\begin{eqnarray}\label{AS_nu}
  \nu =\frac{\psi}{\omega} 							
\end{eqnarray}

While, fixing $\vec{a}_i$, $\vec{a}_j, \nu$, and deriving (\ref{ASDual_for_mu}) by $\mu_k$, we get:
\begin{eqnarray}\label{ASDual_for_mu}
  \frac{\partial D(\vec{\alpha} + \vec{\varepsilon}, \vec{\lambda} + \vec{\mu})}{\partial \mu_k } = \\
  \frac{\partial}{\partial \mu_k} -\frac{1}{4C} (\mu_{k}^{2}S_{k,k}^{-1} + 2 \mu_{k} \nu \vec{y}_i S_{k,i}^{-1} - 2 \mu_{k} \nu \vec{y}_j S_{k,j}^{-1})  = 0
\end{eqnarray}

then:
\begin{eqnarray}\label{AS_mu_k}
  \mu_{k} = \frac{\nu ( \vec{y}_i S_{k,i}^{-1} + \vec{y}_j S_{k,j}^{-1}) }{S_{k,k}^{-1}} 
\end{eqnarray}

These equations, (\ref{AS_nu}, \ref{AS_mu_k}) are used for the updates described in (\ref{SMOS_update_var_alpha_i}, \ref{SMOS_update_var_alpha_j}, and \ref{SMOS_update_var_lambda_k}).

The increment variables are then clipped as follows:

If $(\vec{a}_i + \vec{y}_i  \nu < 0)$ then $\nu = -\vec{y}_i (\text{previous}) \vec{a}_i$

If $(\vec{a}_j - \vec{y}_j  \nu < 0)$ then $\nu =  \vec{y}_j (\text{previous}) \vec{a}_j$

If $(\lambda_k + \mu_k <0)$ then $\lambda_k = 0 $


Let SelectPatterns($i, \mu_k, k$) a procedure that selects all patterns $j$ that after an optimal $\nu$ update gives an increment above a threshold $\beta$, and $\text{dof}$ is the acronym for "dual objective function":

\begin{equation}
  \beta = \delta * (\text{best dof} - \text{previous dof}) + \text{previous dof}
\end{equation}

where $0 < \delta \leq 1$, $\text{best dof}$ is the maximum dual objective function value obtained for all $j \in [1,\dots, l]$, $\text{previous dof}$ is the initial value of the dual objective function without any updates on $\vec{a}_i$ or $\vec{a}_j$. We found that the optimal value of $\delta$ is $1$, so only $1$ pattern is selected for further real update.


While the new $b$ is computed at each iteration with:

\begin{equation}\label{det_b}
b_{\text{new}} \leftarrow \sum_{i = 1 | \xi_i > 0}^{l} \{y_i - [f(\vec{x}_i) - b_{\text{old}}]\} / n
\end{equation}

where $n$ is the number of positive $\xi_i$'s.

We introduced a solver monitor script, that eliminates the solver process whenever it employs more than $120$ seconds to converge to a solution of the generalized quadratic loss SVM problem. This monitor is employed also to stop $\text{SVM}^{\text{light}}$ whenever it takes too much time to converge to a solution.

\subsection{The RTS optimization algorithm}\label{opt3}
A last attempt to solve the model problem is the Representer Theorem with S (RTS), where we consider solutions in the form:

\begin{equation*}
f(\vec{x}) = \sum_{i=1}^{l} \vec{a}_i K(\vec{x}_i,\vec{x}) + b
\end{equation*}

from the Representer Theorem \cite{representer}, with $\vec{a}_i \in {\rm I\!R}$, $i \in {1,\dots,l}$, and $b \in {\rm I\!R}$.
We solve the problem directly in its primal form.

For each variable $\vec{a}_i$ we take a Newton step:

\begin{equation}\label{update_var_alpha_i}
\vec{\alpha}_{i+1} \leftarrow \vec{\alpha}_i - \frac{ \frac{\partial P_{S} }{\partial \vec{a}_i} } {\frac{\partial^{2} P_{S}}{\partial \vec{a}_i^{2}}}
\end{equation}

The determination of $b$ is the same as showed before (\ref{det_b}).
We exit the main loop whenever the problem diverges, or, for 100 consecutive steps, the Newton update is unable to lower the lowest objective function value found.
One advantage of this approach is that it doesn't need the $S$ matrix inversion.
A monitor is used in order to eliminate processes that last more than $120$ seconds.

\subsection{The Deep Learning Framework}\label{DLS}
For a tutorial and survey on Deep Neural Networks you can read \cite{DNN_survey}.
We use Keras 2.3.1 on TensorFlow 2.1.0 back-end in order to obtain Deep Neural Networks that are able to classify a number of input patterns of the order of $1E5$.
The initial shallow network is made of 3 layers: one input layer, one dense and regularized layer, and a sigmoid output layer.
Then we add another dense layer, make algorithms comparisons, and at last, we compare our algorithms with 3 dense layers.
In order to use the generalized quadratic loss we set the batch size to the entire training data-set and write a custom loss that considers pattern correlations.
The $S$ matrix used here is defined in (\ref{S_definition}). Therefore, the custom loss can be written as:





\begin{equation}\label{DL_custom_loss}
  \text{loss} = \vec{o}^{\ t} S \vec{o}
\end{equation}

where $\vec{o}$ is the binary cross-entropy standard loss and $t$ is the transposition operation.

We explore also a regression task with a neural network made of 10 dense and regularized layers.
In this case the custom loss is still (\ref{DL_custom_loss}) but, the output it's the square of the error between the true value and the predicted value, and we consider only the patterns that are present in the current batch in order to build $S$.

\section{Results}\label{results}
We evaluate the results on 2 artificially created data-sets, with different concentration of patterns, which could help in understanding how this loss can support in solving the proposed task. For example, a uniform distribution of patterns and the use of the generalized quadratic loss should not improve the generalization performance w.r.t. a linear loss. While a clustered distribution of patterns should highlight the benefit of the generalized quadratic loss.
The uniform distribution is generated with 800 random points uniformly distributed in the 3 axes, with a random target value randomly chosen in $\{-1,+1\}$.
The normal distribution is generated with 800 random points, spread with a normal distribution with $0$ mean and standard deviation equals to $3$ in the 3 axes, with a random target value randomly chosen in $\{-1,+1\}$.
The calibration procedure is described below.

\begin{figure}
  \caption{A uniform distribution and a normal distribution in order to test the benefits of the generalized quadratic loss.}
  \begin{center}
    \includegraphics[scale=0.35]{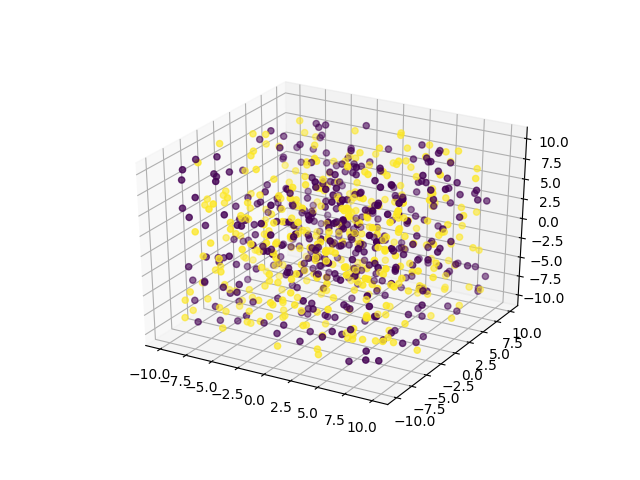} \label{plots}
     \includegraphics[scale=0.35]{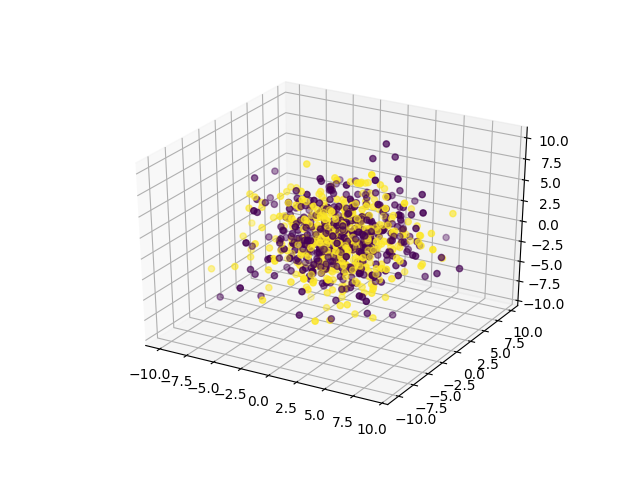}
  \end{center}
\end{figure}

We show the two distributions on figure (\ref{plots}).

For the artificial data-sets, we don't report the results of the SMOS algorithm for a time constraint.

We also evaluate the algorithms on 5 UCI binary classification data-sets: Breast Cancer Wisconsin (Original), Haberman's Survival, Ionosphere, Connectionist Bench (Sonar, Mines vs. Rocks), Twitter Absolute Sigma 500 \footnote{https://archive.ics.uci.edu/ml/datasets.php}, and an UCI regression data-set: YearPredictionMSD.
The first 4 data-sets are used for SVM evaluation, while the Twitter and Year data-set are most suited for Multi-layer Neural Network Learning.
Each data-set is shuffled and divided in a training, validation, and test sets.
The Year data-set is an exception, because it is not shuffled.


    
The training dimensions (number of patterns $\times$ number of features), and the validation and test number of patterns are reported on table (\ref{tab:datasets}).

\begin{table}
	\caption{Data-sets splits dimensions (the number at the right of the $\times$ symbol is the number of features $d$).While the number between parentheses is the number of patterns that are not repeated in the training set.} \label{tab:datasets}
	\begin{center}
		\begin{tabular}{l|l|l|l}
			\textbf{DS Name}  &\textbf{training} & \textbf{validation} &\textbf{test} \\
                        Uniform           & 550    $\times$ 3         &    150 &     100          \\
                        Normal            & 550    $\times$ 3         &    150 &     100          \\
			Breast         	  & 400    $\times$ 9  (281)  &    150 &     149          \\
			Haberman          & 240    $\times$ 3  (227)  &     30 &      36          \\
			Iono              & 260    $\times$ 34 (260)  &     50 &      41          \\
			Sonar             & 130    $\times$ 60 (130)  &     40 &      38          \\
                        Twit              & 5000   $\times$ 77        &  15000 &  120000          \\
                        Year              & 150000 $\times$ 90        & 313000 &   51000          \\
		\end{tabular}
	\end{center}
\end{table}

SMOS and RTS procedures are calibrated on three hyper-parameters: $C$, $\gamma_K$, and $\gamma_S$ on a grid of $6 \times 6 \times 6$ with steps of powers of $10$ starting from $(0.001, 0.001, 0.001)$ to $(100, 100, 100)$.
For each data-set, the set of hyper-parameters that gave the best result, in term of $F1$ measure on the validation set, is saved together with the best dual variables $\vec{\alpha}$ and best $b$. Then these saved variables are used to evaluate the test set, which outputs the final F1 performance on that data-set.
We implement a SMO-like algorithm inspired by \cite{SMO-like} and \cite{SMO} to solve problem (\ref{dual_S}), and we name it SMOS.
Besides we propose another simpler algorithm that works only in the primal problem (\ref{opt3}).
At last we front the problem with an algorithm based on an objective function made only by the error term with $S$ and no regularization term.
We call this algorithm simply S.
This procedure is calibrated on two hyper-parameters: $\gamma_K$ and $\gamma_S$ on a grid of $10 \times 10$ with steps of powers of $10$ starting from $(1E-5, 1E-5)$ to $(10000, 10000)$.

The F1 scores on each data-set are reported in table {\ref{tab:resultsS}}, together with the F1 score of SMOS, RTS, S, and the well known $\text{SVM}^{\text{light}}$ software \cite{Joachims}.

\begin{table}
	\caption{Results comparison with SMOS, RTS, S, and $\text{SVM}^{\text{light}}$ with standard linear loss.} \label{tab:resultsS}
	\begin{center}
		\begin{tabular}{lllllll}
			\textbf{Algorithm} & \textbf{DS Name}  &\textbf{C} & \textbf{$\gamma_K$} &\textbf{$\gamma_S$} &\textbf{F1}& \textbf{Time}  \\
			\hline \\
                        RTS      & Uniform         &  100 & 10 & 10 & 0.589928 & 78m24s092ms \\
                        S        & Uniform         &      & 1E-5 & 0.01 & \textbf{0.620690} & 38m50s996ms \\
                        $\text{SVM}^{\text{light}}$   & Uniform & 1E-05 & 1E-05 & & \textbf{0.620690} & 0m43s178ms \\ \cline{1-7}
                        RTS      & Normal          & 0.01 & 100 & 0.001 & \textbf{0.722581} & 51m29s542ms \\
                        S        & Normal          &      & 100 & 1E-05 & \textbf{0.722581} & 40m57s976ms \\
                        $\text{SVM}^{\text{light}}$   & Normal  & 10000 & 0.1 & & 0.519206 & 8m29s650ms \\ \cline{1-7}
			SMOS     &Breast         & 0.1    & 0.01  & 1 & 0.961538     & 5h28m8s453ms      \\
			RTS      &Breast 	 & 0.1   & 100 & 0.001   & 0.938053  &   62m41s721ms      \\
                        S        &Breast         &               &0.001   & 1E-5  & 0.971429 & 6m18s446ms\\
			$\text{SVM}^{\text{light}}$ &Breast         & 1   & 0.0   &       & \textbf{0.971462} & 2m08s630ms\\ \cline{1-7}
			SMOS     &Haberman       & 0.1 & 0.001 & 10    & 0.885246  & 2h27m9s594ms\\
			RTS      &Haberman  	 & 0.1 & 0.1    & 1  & 0.857143 & 75m51s646ms\\
                        S        &Haberman       &              & 0.001 & 0.1 & 0.896552  & 4m6s346ms  \\
			$\text{SVM}^{\text{light}}$ &Haberman       & 1   & 0.001 &   & 0.857143 &     0m18s850ms          \\\cline{1-7}
			SMOS     &Iono           & 1   &  0.1  & 100   & \textbf{0.958333}  &   6h22m10s739ms                   \\
                        RTS      &Iono		 & 100 &  1    & 1     & 0.893617           & 47m38s634ms    \\
                        S        &Iono           &     & 0.1   & 100   & \textbf{0.958333}                   & 6m59s161ms \\
			$\text{SVM}^{\text{light}}$ &Iono           & 10  & 0.1   &            & 0.9583&   6m15s790s          \\\cline{1-7}
			SMOS     &Sonar          & 1     & 1     & 10                       & 0.800000  & 3h39m10s539ms        \\
			RTS      &Sonar          & 100   & 1     & 1                        & 0.742857   & 47m50s459ms\\
                        S        &Sonar          &       & 0.1   & 0.1                      & 0.820513 &  8m50s875ms   \\
                        $\text{SVM}^{\text{light}}$ &Sonar          & 10  & 1     &            & 0.800000   & 7m07s650ms  \\
		\end{tabular}
	\end{center}
\end{table}

We report in table (\ref{tab:results_fold}) some experiments with a $10$-fold cross-validation scheme, as described in \cite{cross_validation}, in order to give a greater significance level. We created 10 subsets of the training sets. Each algorithm is run $10$ times, one time for each $k$ sub training data-set. We consider the mean and the standard deviation of each resulting test F1 scores. The parameter grid explored is the same as defined before.
The results with the SMOS method are not reported due to a time constraint.

\begin{table}
	\caption{Results comparison with RTS, S, and $\text{SVM}^{\text{light}}$ with standard linear loss with a 10-fold cross-validation procedure.} \label{tab:results_fold}
	\begin{center}
		\begin{tabular}{lllll}
			\textbf{Algorithm} & \textbf{DS Name}  & \textbf{F1} & $\sigma$ & \textbf{Time}  \\
			\hline \\
                        RTS      & Uniform        & 0.607557 & 0.013028 & 45h06m    \\ 
                        S        & Uniform        & \textbf{0.62069}  & \textbf{1.110223E-16} & 4h45m08s  \\
			$\text{SVM}^{\text{light}}$  & Uniform     & 0.611796 & 0.811098 & 13m56s54ms \\\cline{1-5}
                        RTS      & Normal         & \textbf{0.719074} & 0.0156605 & 38h39m \\
                        S        & Normal         & 0.717949 & \textbf{0.0} & 5h04m20s \\
			$\text{SVM}^{\text{light}}$  &Normal  & 0.534307 & 7.869634 & 1h01m23s \\\cline{1-5}
                        RTS      & Breast         & \textbf{0.975627} & \textbf{0.006172} & 31h57m \\ 
                        S        & Breast         & 0.946792 & 0.012488 & 4h45m350ms \\
			$\text{SVM}^{\text{light}}$ & Breast    & 0.967170 & 0.677366 & 12m15s672ms  \\\cline{1-5}
                        RTS      & Haberman      & \textbf{0.938325} & 0.033040 & 33h47m  \\
                        S        & Haberman      & 0.865720 & \textbf{0.018494} & 2h3m29s888ms  \\
			$\text{SVM}^{\text{light}}$ &Haberman  & 0.858132 & 2.340863 & 3m14s972ms \\\cline{1-5}
                        RTS      & Iono          & 0.961435 & 0.016616 & 29h13m     \\
                        S        &Iono           & \textbf{0.971863} & \textbf{0.016387} & 5h17m54s   \\
			$\text{SVM}^{\text{light}}$ &Iono      & 0.915835 & 1.784254 & 44m42s632ms\\\cline{1-5}
                        RTS      & Sonar         & \textbf{0.934440} & \textbf{0.026761} &  22h55m \\
                        S        &Sonar          & 0.800656 & 0.059841 & 5h15m18s655ms \\
                        $\text{SVM}^{\text{light}}$ &Sonar     & 0.843513 & 0.794849 & 1h10m2ms953s \\
		\end{tabular}
	\end{center}
\end{table}

Moreover, we describe some experiments in table (\ref{tab:resultsSMKL}) obtained with a $K$ and $S$ matrices generated with multiple kernels. We call this algorithm $\text{RTS}_{\text{MKL}}$.

\begin{equation}
  K_{MKL}(\vec{x_i}, \vec{x_j}) = \sum_{i=1}^{nK} b_i * K_i(\vec{x_i}, \vec{x_j}) 
\end{equation}

\begin{equation}
  S_{MKL}(\vec{x_i}, \vec{x_j}) = \sum_{i=1}^{nS} c_i * S_i(\vec{x_i}, \vec{x_j}) 
\end{equation}

where $b_i$ and $c_i$ are heuristically determined as in \cite{heuristic_MKL}, the kernels $K_i()$ and $S_i()$ are defined as RBF kernels with different $\gamma$ (for the definition see (\ref{S_definition})), and we set the maximum number of $nK=10$ and the maximum number of $nS=10$.
We report also some experiments with an algorithm based on the Representer Theorem with a linear loss and MKL, in order to try to understand if the benefits on the generalization performance is given by the MKL technique or the use of $S$. We call this algorithm $\text{RT}_{\text{MKL}}$.
We add that a monitor is adopted in order to eliminate $\text{SVM}^{\text{light}}$ processes that take more than $120$ seconds to converge, while no monitor is used for the RTS with MKL algorithm.
In the last column, we report the calibration time plus the test evaluation time.

\begin{table}
	\caption{Results comparison with RT with MKL and linear loss, RTS with MKL and S, and $\text{SVM}^{\text{light}}$ with standard linear loss.} \label{tab:resultsSMKL}
	\begin{center}
		\begin{tabular}{llllllll}
			\textbf{Algorithm} & \textbf{DS Name}  &\textbf{C} & $\mathbf{\gamma_K}$ & \textbf{best nKK} & \textbf{best nKS} &\textbf{F1}        & \textbf{Time}  \\
			\hline                                                                                                       \\
                        $\text{RT}_{\text{MKL}}$  &     Breast   & 10   &     &  1           &                  & 0.953271          & 6m44,947s     \\            
			$\text{RTS}_{\text{MKL}}$ &     Breast   & 0.1   &    &  2            & 1                 &  \textbf{0.972477}         & 61m6,436s     \\
			$\text{SVM}^{\text{light}}$ & Breast         & 1   & 0.1 &   &                & 0.971462 &  2m08s630ms     \\ \cline{1-7}
                        $\text{RT}_{\text{MKL}}$    & Haberman   & 0.001   &      &   3          &                  &  0.857143         & 0m16,233s      \\
			$\text{RTS}_{\text{MKL}}$   & Haberman & 0.1      & 0.01      &  3        &  9                & \textbf{0.896552} & 24m55,662s      \\
			$\text{SVM}^{\text{light}}$ & Haberman       & 1   & 0.001 &  &               & 0.857143          & 0m18s800ms      \\\cline{1-7}
                        $\text{RT}_{\text{MKL}}$    & Iono   &  0.01  &      & 2            &                  &  0.836364         &  0m23,713s    \\
			$\text{RTS}_{\text{MKL}}$  &    Iono           & 100  &       & 2          &  7               & 0.938776          & 35m12,091s       \\
			$\text{SVM}^{\text{light}}$ & Iono           & 10 & 0.1    &   &              & \textbf{0.9583}   & 0m50,612s      \\\cline{1-7}
                        $\text{RT}_{\text{MKL}}$    & Sonar   & 0.01   &      & 1            &                  & \textbf{0.800000}       & 0m11,134s     \\
                        $\text{RTS}_{\text{MKL}}$  &    Sonar          &  1        &  & 2  & 1             & 0.717949 & 14m19,842s      \\
			$\text{SVM}^{\text{light}}$ & Sonar          & 10 & 1      &    &             & \textbf{0.800000} & 1m18,435s      \\
		\end{tabular}
	\end{center}
\end{table}

We report some experiments with different shapes of Neural Networks applied to the UCI Buzz in Social Media-Twitter data-set, in table (\ref{tab:resultsTwitter}), where the algorithm is described in section (\ref{DLS}).
Hence we name SNN the standard Shallow multi-layer Neural Network algorithm, while we name GQLSNN the Shallow Neural Network with a generalized quadratic loss.
We make a grid search to find the optimal hyper-parameters. We start with one dense layer made of $5$ nodes and then we increase this size until we reach $25$ nodes, following steps of $5$.
The same sequence is applied with the case with 2 and 3 dense and regularized layers.
For the GQLSNN algorithm we add another hyper-parameter, that is $\gamma_S$. The line search on this hyper-parameter spans through $[1E-5, \dots, 0.1]$ with steps of powers of $10$.

Each network run comprises a training procedure of $1000$ epochs, with a batch size of $5000$ patterns.
We selected the hyper-parameters that perform at best from the whole grid, on the validation data-set.
The test F1 score reported is determined by the evaluation of the best validation model on the test data-set.

\begin{table}
	\caption{Results comparison with SNN, and GQLSNN exploiting $S$ with the Twitter data-set, with the optimal hyper-parameters: the number of nodes for layers number $2$ to $4$, the optimal $\gamma_S$ selected by the calibration procedure, the training, validation and test time.} \label{tab:resultsTwitter}
	\begin{center}
		\begin{tabular}{l|l|l|l|l|l|l|l}
		\textbf{Algorithm}  & \textbf{nNL2} & \textbf{nNL3} &\textbf{nNL4} &$\mathbf{\gamma_S}$ & \textbf{test F1}& \textbf{Time}\\
                        SNN     & 15     &     &       &           &     0.88805829          & 1m5s274ms     \\
                        GQLSNN  & 25     &     &       &  0.1      &     \textbf{0.90734076} & 59m47s075ms   \\
                        SNN     & 5      & 15  &       &           &     0.9005088           & 10m40s482ms   \\
			GQLSNN  & 5      & 10  &       & 0.0001    &     \textbf{0.9118929}  & 5h21m20s065ms \\
                        SNN     & 25     & 20  & 10    &           &     0.9080214           & 28m9s709ms    \\
			GQLSNN  & 10     & 10  & 20    & 0.1       &     \textbf{0.91080284} &  26h53m59s    \\
		\end{tabular}
	\end{center}
\end{table}

For the Year data-set, described in details in \cite{MillionSongs_dataset}, we trained a deeper network made of 10 dense and regularized layers, with 100 nodes for each layer. The batch size is 1000 and the number of epochs is 400.
The standard Deep Neural Network setting is named DNN.
For the GQLDNN algorithm the $\gamma_S$ was calibrated with a line search that spans through $[1E-5, \dots, 0.1]$ with steps of powers of $10$.
We measure the Mean Square Error, and the performance is reported in table (\ref{tab:resultsYear}).
Every Neural Network we employ uses the Adam optimizer
\footnote{This software runs on a Intel(R) Core(TM) i7-6700 CPU @ 3.40GHz with 32.084 MB of RAM, 32.084 MB of swap space, and a SSD of 512 GB.}.

\begin{table}
	\caption{Results comparison with DNN, and GQLDNN exploiting $S$, with the optimal $\gamma_S$ selected by the calibration procedure.} \label{tab:resultsYear}
	\begin{center}
		\begin{tabular}{l|l|l|l}
		\textbf{Algorithm}  & $\mathbf{\gamma_S}$ & \textbf{test MSE} & \textbf{Time} \\
                        DNN     &           &             129.3600     & 10m17s420ms          \\
                        GQLDNN  &    0.0001 &     \textbf{114.95802}   & 4h52s02ms            \\
		\end{tabular}
	\end{center}
\end{table}

\section{Conclusions}\label{conclusions}
We can try other S matrices, in order to furtherly generalize the results. 
Another improvement could be to tune the MKL coefficients with an optimization procedure as suggested by \cite{optimized_MKL_coef}.
In addition, Support Vector Regression and multi-class classification with a generalized quadratic loss could be investigated.



The results obtained with the Twitter data-set, a Neural Networks with 2, 3, 4 layers, and a generalized quadratic loss, are encouraging.

For the regression setting we tried a deeper neural network on a larger data-set (UCI Year) and we found that the generalized quadratic loss performed in a similar manner than standard sum of the errors loss' square.

We still have to realize if the performances of the GQL for the Shallow and Deep Neural Networks are due to the greater number of trials needed to tune the S matrix, or to a real effect on the generalization ability of the algorithm induced by the use of the GQL.
We made some preliminary experiments to establish this, but it is too premature to make an assertion.

\subsubsection*{Acknowledgments}
I would like to express my gratitude to Giovanna Zamara, Fabrizio Romano, Fabio Aiolli, Alessio Micheli, Ralf Herbrich, Alex Smola, Alessandro Sperduti for their insightful suggestions.

%
%
%
%

\end{document}